\PassOptionsToPackage{dvipsnames}{xcolor}
\documentclass{article}

\usepackage[final]{colm2026_conference}

\usepackage[T1]{fontenc}
\usepackage{microtype}
\usepackage{amsmath,amssymb}
\usepackage{booktabs}
\usepackage{multirow}
\usepackage{tabularx}
\usepackage{graphicx}
\usepackage{subcaption}
\usepackage{enumitem}
\usepackage{lineno}
\usepackage{float}

\usepackage{listings}
\usepackage[dvipsnames]{xcolor}
\usepackage{hyperref}
\definecolor{darkblue}{rgb}{0, 0, 0.5}
\hypersetup{colorlinks=true, citecolor=darkblue, linkcolor=darkblue, urlcolor=darkblue}

\newif\ifreviewmode
\reviewmodefalse  

\ifreviewmode
  \newcommand{\fix}[1]{\textcolor{red}{\textbf{[FIX: #1]}}}
  \newcommand{\note}[1]{\textcolor{blue}{\textbf{[NOTE: #1]}}}
  \newcommand{\todo}[1]{\textcolor{orange}{\textbf{[TODO: #1]}}}
  
\else
  \newcommand{\fix}[1]{}
  \newcommand{\note}[1]{}
  \newcommand{\todo}[1]{}
  
\fi

\usepackage{pifont}
\newcommand{\cmark}{\ding{51}}    
\newcommand{\xmark}{\ding{55}}    

\newcommand{\benchguard}{\textsc{BenchGuard}}
\newcommand{\cat}[1]{\textsc{#1}}       
\newcommand{\subcat}[1]{\texttt{#1}}    
\newcommand{\sev}[1]{\textbf{#1}}       
\newcommand{\code}[1]{\texttt{#1}}

\title{Who Guards the Benchmarks? Automated Auditing of LLM Agent Benchmarks}

\author{
  \begin{tabular}{ccc}
  Xinming Tu$^{1,2,*}$ & Tianze Wang$^{1}$ & Yingzhou (Minta) Lu$^{2}$ \\
  Kexin Huang$^{2}$ & Yuanhao Qu$^{2,*}$ & Sara Mostafavi$^{1,3,*}$ \\
  \end{tabular} \\
  \\
  $^{1}$Paul G. Allen School of Computer Science \& Engineering, \\
  \hphantom{$^{1}$}University of Washington, Seattle, WA, USA \\
  $^{2}$Phylo, Inc., South San Francisco, CA, USA \\
  $^{3}$Genentech, Inc., South San Francisco, CA, USA \\
  {\small $^{*}$Correspondence: \texttt{tuxm@cs.washington.edu},
  \texttt{yuanhao@phylobio.com}, \texttt{saramos@cs.washington.edu}} \\
  {\small \textbf{Code:} \url{https://github.com/XinmingTu/BenchGuard}}
}

\usepackage[most]{tcolorbox}

\definecolor{caseBg}{HTML}{FFF5F5}    
\definecolor{caseFrame}{HTML}{C53030} 
\definecolor{caseHl}{HTML}{FBB6B6}    

\newcommand{\hlcode}[1]{{\setlength{\fboxsep}{1.5pt}\colorbox{caseHl}{\texttt{#1}}}}
\newcommand{\hltext}[1]{{\setlength{\fboxsep}{1.5pt}\colorbox{caseHl}{\textbf{#1}}}}

\definecolor{promptBg}{HTML}{F7FAFC}
\definecolor{promptFrame}{HTML}{2B6CB0}

\newtcolorbox{promptbox}[2][]{
  enhanced, breakable,
  colback=promptBg,
  colframe=promptFrame,
  boxrule=0.6pt,
  arc=1mm,
  left=3mm, right=3mm, top=1mm, bottom=1.5mm,
  fontupper=\footnotesize,
  before upper=\linespread{0.88}\selectfont,
  halign=flush left,
  fonttitle=\bfseries\footnotesize,
  coltitle=white,
  colbacktitle=promptFrame,
  title={#2},
  before skip=8pt plus 2pt,
  after skip=8pt plus 2pt,
  #1
}

\newtcolorbox{casebox}[3][]{
  enhanced,
  colback=caseBg,
  colframe=caseFrame,
  boxrule=0.6pt,
  arc=1mm,
  left=3mm, right=3mm, top=1mm, bottom=1.5mm,
  fontupper=\footnotesize,
  before upper=\linespread{0.88}\selectfont,
  halign=flush left,
  fonttitle=\bfseries\footnotesize,
  coltitle=white,
  colbacktitle=caseFrame,
  title={#2\hfill#3},
  before skip=8pt plus 2pt,
  after skip=8pt plus 2pt,
  #1
}

\definecolor{boundBg}{HTML}{F0FDFA}
\definecolor{boundFrame}{HTML}{0D9488}
\definecolor{boundHl}{HTML}{99F6E4}

\newcommand{\bhlcode}[1]{{\setlength{\fboxsep}{1.5pt}\colorbox{boundHl}{\texttt{#1}}}}
\newcommand{\bhltext}[1]{{\setlength{\fboxsep}{1.5pt}\colorbox{boundHl}{\textbf{#1}}}}

\newtcolorbox{boundarybox}[3][]{
  enhanced,
  colback=boundBg,
  colframe=boundFrame,
  boxrule=0.6pt,
  arc=1mm,
  left=3mm, right=3mm, top=1mm, bottom=1.5mm,
  fontupper=\footnotesize,
  before upper=\linespread{0.88}\selectfont,
  halign=flush left,
  fonttitle=\bfseries\footnotesize,
  coltitle=white,
  colbacktitle=boundFrame,
  title={#2\hfill#3},
  before skip=8pt plus 2pt,
  after skip=8pt plus 2pt,
  #1
}

\begin{document}

\ifcolmsubmission
\linenumbers
\fi

\maketitle

\begin{abstract}
As benchmarks grow in complexity, many apparent agent failures are not failures of the agent at all---they are failures of the benchmark itself: broken specifications, implicit assumptions, and rigid evaluation scripts that penalize valid alternative approaches. We propose employing frontier LLMs as systematic auditors of evaluation infrastructure, and realize this vision through \benchguard{}, the first framework explicitly designed for joint cross-artifact auditing of execution-based agent benchmarks. \benchguard{} cross-verifies all benchmark artifacts via structured LLM protocols, optionally incorporating agent solutions or execution traces as additional diagnostic evidence. Deployed on two prominent scientific benchmarks, \benchguard{} identified 12 author-confirmed issues in ScienceAgentBench---including fatal errors rendering tasks unsolvable---and exactly matched 83.3\% of expert-identified issues on the BIXBench Verified-50 subset, catching defects that prior human review missed entirely. A full audit of 50 complex bioinformatics tasks costs under \$15, making automated benchmark auditing a practical and valuable complement to human review. A preliminary native-format audit of ProgramBench further demonstrates cross-format applicability. These findings point toward AI-assisted benchmark development, where frontier models serve not only as subjects of evaluation but as active participants in validating the evaluation infrastructure itself.
\end{abstract}

\section{Introduction}
\label{sec:intro}

As LLM agents tackle increasingly complex tasks---resolving software issues~\citep{jimenez2024swebench}, navigating web environments~\citep{zhou2024webarena}, and automating end-to-end AI research~\citep{lu2026aiscientist}---a growing ecosystem of execution-based benchmarks has emerged to measure these capabilities~\citep{jimenez2024swebench, zhou2024webarena, xie2024osworld, merrill2026terminalbench}.
Unlike static, multiple-choice datasets~\citep{hendrycks2021mmlu}, these benchmarks are tightly coupled evaluation pipelines that comprise natural language instructions, executable ground-truth programs, evaluation scripts, and containerized environments, making the measurement infrastructure itself susceptible to systemic logical flaws.
This fragility is particularly acute in specialized scientific domains~\citep{chen2024scienceagentbench, mitchener2025bixbench}, where evaluation demands deep domain expertise.

Relying on domain experts to construct these benchmarks is necessary, and the limitations of current benchmarking methodology are well recognized~\citep{bowman2020fixbenchmarking}, yet recent benchmark-repair efforts show that even widely used, human-validated evaluations contain enough noise to distort capability measurements---pervasive label errors have been documented across flagship static benchmarks~\citep{gema2024mmlu, northcutt2021pervasive}.
In execution-based settings, the problem is compounded: SWE-bench Verified was constructed specifically to filter unreliable tasks, yet subsequent analysis found residual issues even in this curated subset~\citep{openai2024swebenchverified, openai2026nolonger}. This noise is uniquely difficult to detect because correctness emerges from interactions among instructions, ground-truth programs, evaluation scripts, and environments---not from any single label.
We interpret many of these persistent errors through the lens of \emph{solution fixation}~\citep{jansson1991designfixation} and the curse of knowledge~\citep{camerer1989curse}: benchmark creators anchor to their own implementations and unconsciously assume that implicit choices are self-evident, leaving instructions underspecified~\citep{gema2024mmlu} and evaluation suites insufficient or misaligned~\citep{liu2023evalplus, openai2026nolonger}---a pattern consistent with pervasive annotation errors in static benchmarks~\citep{northcutt2021pervasive}, but harder to detect when spread across interacting artifacts.

To break this bottleneck, we propose employing frontier LLMs not merely as subjects of evaluation or output judges~\citep{zheng2023mtbench, li2024llmasjudge}, but as systematic auditors of the evaluation infrastructure itself---a direction supported by evidence that LLM critics can identify defects that human oversight missed~\citep{mcaleese2024llmcritics}.
We introduce \benchguard{}, the first framework explicitly designed for joint cross-artifact auditing of execution-based agent benchmarks.
Its methodological primitive is a single structured cross-artifact prompt: it has no learned component or inter-task reasoning, so the contribution is primarily conceptual, empirical, and workflow-oriented rather than a new learning algorithm.
While prior work has produced verified or revised benchmark variants through re-annotation and model-assisted auditing for static or question-answering settings~\citep{gema2024mmlu, zhai2026hleverified}, and through expert re-annotation for execution-based settings such as SWE-bench Verified~\citep{openai2024swebenchverified}, no existing framework systematically audits the coupled artifact stack of execution-based benchmarks.
The framework cross-references all benchmark artifacts via structured LLM protocols, optionally incorporating agent solutions or execution traces as benchmark-dependent diagnostic evidence.
We also introduce an empirically grounded taxonomy of 4 categories and 14 subcategories for classifying benchmark defects.
We scope the framework to task-oriented, execution-based benchmarks (Section~\ref{sec:overview}).

Deployed on two prominent agent benchmarks, \benchguard{} exposed the fragility of current human-reviewed evaluations.
On ScienceAgentBench, despite the benchmark undergoing multiple rounds of manual validation by annotators and subject matter experts, we identified 12 defects---all subsequently confirmed by the original authors.
On BIXBench, which underwent independent review prior to publication, a five-model \benchguard{} ensemble exactly matched 83.3\% of expert-identified issues on the Verified-50 subset, while also surfacing strong possible artifact inconsistencies that prior manual review had missed.
These results are consistent with growing evidence that follow-up audits continue to uncover benchmark flaws even after careful curation~\citep{gema2024mmlu, siddiq2024faultinstars, openai2026nolonger}, but reveal a new dimension of the problem specific to execution-based benchmarks: tightly coupled software artifacts create failure modes invisible to dataset-only auditing, suggesting that automated, AI-assisted auditing is a valuable complement to human review in the benchmark lifecycle.
Cases~1--2 illustrate two such failures discovered in ScienceAgentBench, where pink highlights mark the cross-artifact discrepancy:

\begin{tcbraster}[raster columns=2, raster equal height=rows,
  raster column skip=2.5mm, raster row skip=3mm]

\begin{casebox}{Case 1: Wrong Input File}{\subcat{INST}}
\textbf{Instruction:}\\
\textit{``...perform the analysis using \hltext{ecg\_1000hz.csv}...''}\\[0.3em]
\textbf{Reference Solution:}\\
\texttt{pd.read\_csv(}\\
\texttt{~~'\hlcode{bio\_eventrelated\_100hz.csv}')}\\[0.3em]
{\color{caseFrame}\textbf{Defect:}} Instruction points to the wrong file; the reference solution uses a different one.
\end{casebox}
\begin{casebox}{Case 2: SMILES vs.\ Drug Names}{\subcat{EVAL}}
\textbf{Instruction:}\\
\textit{``...save ordered list of drugs..., with \hltext{one SMILES per line}.''}\\[0.3em]
\textbf{Evaluation Script:}\\
\texttt{if \hlcode{"Foscarnet"} in pred[i]:}\\[0.3em]
{\color{caseFrame}\textbf{Defect:}} Evaluator checks for drug names, contradicting the SMILES format the instruction requests.
\end{casebox}

\end{tcbraster}

\noindent Both defects passed multiple rounds of human review and were subsequently confirmed by the original benchmark authors.
Such cross-artifact mismatches are invisible to single-component review but precisely what automated auditing catches systematically.

Our main contributions are as follows:
\begin{enumerate}[leftmargin=*,itemsep=2pt]
  \item \textbf{AI-Assisted Evaluation Paradigm and Error Taxonomy.} We characterize cross-artifact consistency failures in agent benchmarks (motivated by solution fixation as an explanatory lens) and introduce a 4-category, 14-subcategory error taxonomy grounded in the constituent artifacts of execution-based benchmarks and validated through confirmed defects across production benchmarks.
  \item \textbf{The \benchguard{} Framework.} We present the first framework explicitly designed for joint cross-artifact auditing of execution-based agent benchmarks, capable of auditing 50 complex tasks with five frontier models for under \$15. Code and auditing skills are available at \url{https://github.com/XinmingTu/BenchGuard}.
  \item \textbf{Empirical Validation.} We uncover author-confirmed, solve-rate-altering defects in ScienceAgentBench, demonstrate expert-revision alignment on BIXBench, and provide a preliminary native-format extension to ProgramBench.
\end{enumerate}

We survey related work in Section~\ref{sec:related}, present the framework in Section~\ref{sec:method}, and report empirical findings in Sections~\ref{sec:setup}--\ref{sec:results}.

\begin{figure}[htb]
\centering
\includegraphics[width=0.84\textwidth]{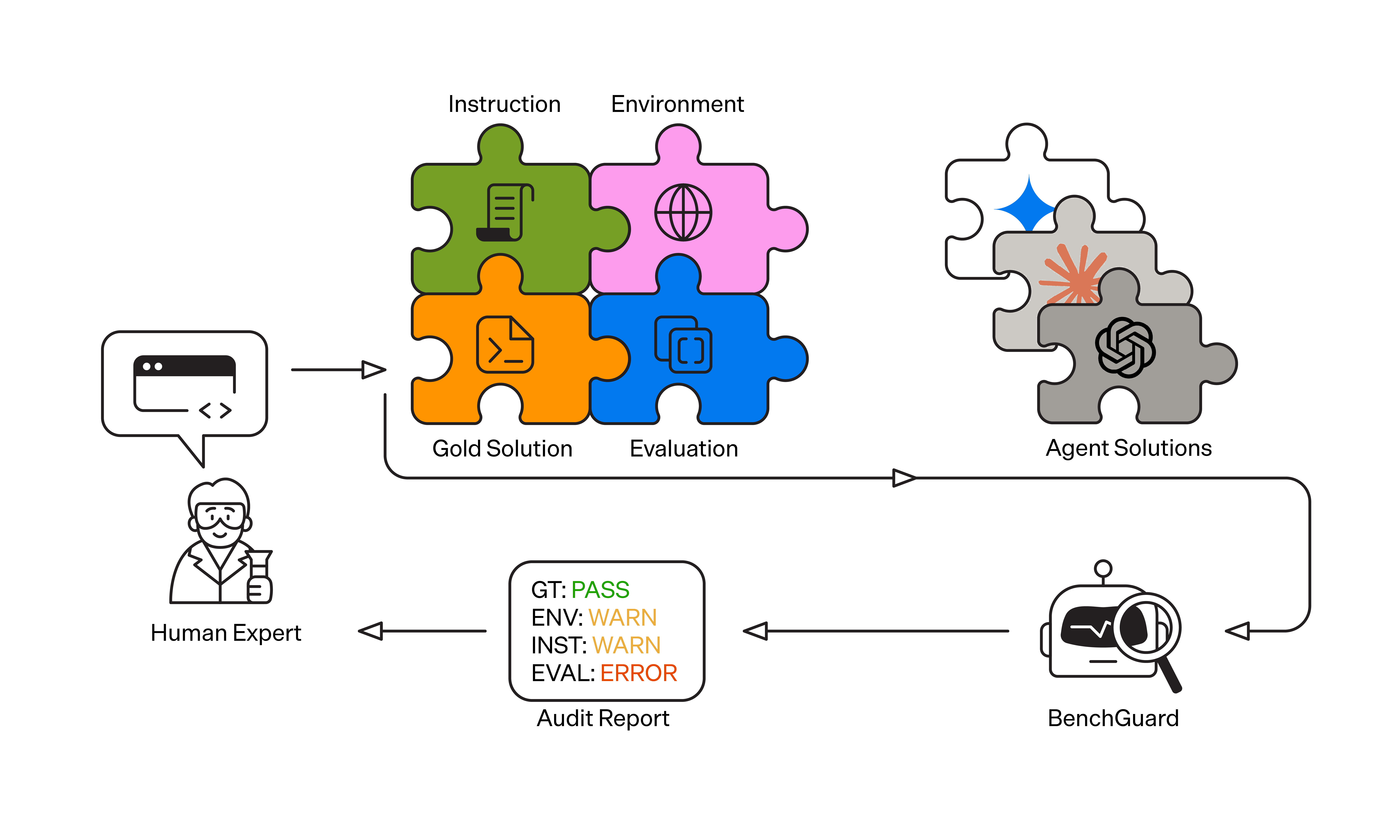}
\caption{Overview of the coupled artifact stack audited by \benchguard{}. Each task is defined jointly by four interlocking components: the natural-language instruction, the environment, the reference gold solution, and the evaluation logic. \benchguard{} treats benchmark auditing as a cross-artifact consistency problem: misalignment among any of these pieces can produce benchmark defects, which are surfaced as structured findings for expert review.}
\label{fig:architecture}
\end{figure}

\section{Related Work}
\label{sec:related}

\benchguard{} is most closely connected to two threads: (i)~post-hoc auditing and ``verified'' revisions of benchmarks, and (ii)~evaluation methodology for execution-based agent benchmarks, where correctness is implemented through coupled executable artifacts rather than static labels.

\subsection{Benchmark Auditing and Verified Revisions}

Benchmark integrity has become a first-order concern as leaderboards saturate and evaluation artifacts are reused at scale. In static QA benchmarks, careful audits have shown that seemingly small annotation or specification errors can materially alter measured performance, motivating verified subsets and re-annotation efforts such as MMLU-Redux~\citep{gema2024mmlu}. Similar dynamics are emerging in newly proposed expert-level evaluations: Humanity's Last Exam (HLE) was introduced to raise the ceiling, yet it quickly attracted follow-up auditing and structured verification efforts, culminating in revision pipelines such as HLE-Verified~\citep{cais2026hle,zhai2026hleverified}. In mathematical reasoning, systematic auditing of the Omni-MATH benchmark revealed dataset-induced noise that necessitated manual revision~\citep{ballon2026smarterjudge}. In the code domain, quality assessments have likewise highlighted that benchmark artifacts beyond labels---including prompts and dataset construction choices---can systematically skew conclusions~\citep{siddiq2024faultinstars}.
In execution-based settings, SWE-bench Verified~\citep{openai2024swebenchverified} adopted human re-annotation to filter unreliable tasks, yet subsequent evidence suggests that even this verified subset contains residual issues~\citep{openai2026nolonger}.

These efforts establish an important lesson: benchmark quality is not guaranteed, and verification is often a necessary second step. However, most prior auditing pipelines still treat the evaluation oracle as relatively self-contained (a label, rubric, or a single checker). Execution-based agent benchmarks break this assumption: correctness is realized through interactions among instructions, ground-truth programs, evaluation logic, and runtime environments, creating failure modes that cannot be captured by dataset-only auditing.

\subsection{Execution-Based Agent Benchmarks and Evaluation Reliability}

Agent benchmarks increasingly assess long-horizon problem solving in executable environments, including repository-level software engineering~\citep{jimenez2024swebench}, scientific analysis workflows~\citep{chen2024scienceagentbench}, and research-style domains such as bioinformatics~\citep{mitchener2025bixbench}. A defining property of these settings is that each task is a coupled system: natural-language specifications are operationalized through evaluation harnesses (scripts, tests, and sometimes model judges) under specific environment assumptions. As a result, mis-measurement can arise from subtle specification--implementation mismatches even when each component appears reasonable in isolation.

Human verification is a pragmatic remedy and has been adopted in prominent benchmarks (e.g., SWE-bench Verified)~\citep{openai2024swebenchverified}. Yet recent evidence suggests that one-off review is not sufficient as a lifecycle strategy: test suites can still reject functionally correct solutions, and contamination and evaluator limitations can increasingly dominate reported scores, leading to shifts away from previously ``verified'' benchmarks for frontier evaluation~\citep{openai2026nolonger}. More broadly, the growing use of model judges in evaluation introduces another source of measurement error: judge reliability can itself become a bottleneck as models advance~\citep{zheng2023mtbench,kim2024prometheus2,ballon2026smarterjudge}.

Closest to our setting, AgentSuite/COBA develops an issue taxonomy and automated pipeline for filtering flawed agent-evaluation tasks~\citep{suh2026agentsuite}, while concurrent ELT-Bench-Verified combines LLM root-cause analysis with human validation for one execution-based data-engineering benchmark~\citep{zanoli2026eltbench}. \benchguard{} instead jointly localizes inconsistencies across instructions, gold artifacts, evaluators, and environments and proposes repairs for expert adjudication. Complementary SWE-bench studies audit candidate patches or augment tests, showing that incomplete tests can accept behaviorally incorrect solutions and alter reported scores~\citep{wang2025solvedissues,yu2025utboost}; our focus is pre-release auditing of benchmark-owned artifacts.

\section{Method}
\label{sec:method}

\subsection{Overview}
\label{sec:overview}

\benchguard{} takes as input the four constituent artifacts of an execution-based benchmark task: the natural-language \emph{instruction}, the \emph{ground-truth program} (reference solution), the \emph{evaluation script}, and the \emph{environment configuration}.
It produces a structured set of findings, each annotated with an error category from our taxonomy (Section~\ref{sec:taxonomy}), a severity level, a confidence score, and supporting evidence with source-level citations.

The core \emph{definition-level audit} (Section~\ref{sec:definition}) performs LLM-driven cross-verification of the four artifacts, supplemented by deterministic static checks, to detect logical inconsistencies visible from the benchmark specification alone.
When agent solutions or execution traces are available, they can be appended to the audit context as additional diagnostic evidence (Section~\ref{sec:execution}).

Concretely, the reference pipeline ingests a standardized format compatible with Harbor~\citep{harbor2026} (Appendix~\ref{app:ingestion}), constructs a verification context for each task bundling all available artifacts, and dispatches it to LLM verification protocols and deterministic checks running in parallel. The skill-based workflow is format-agnostic: a coding agent discovers the four artifact classes in a benchmark's native repository and performs any needed adaptation or conversion, so benchmark authors need not manually reformat tasks into Harbor.
Outputs are aggregated into a structured report with per-task cost tracking.
Rather than acting as a fully autonomous judge, \benchguard{} surfaces high-confidence findings with supporting evidence for expert adjudication, establishing a human--AI collaborative workflow.

We scope \benchguard{} to \emph{task-oriented, execution-based} agent benchmarks: settings where each task has a well-defined input, a target output, and verification logic (deterministic scripts or structured judges, including model-based evaluators with reproducible protocols).
It excludes open-ended conversational agents, interactive tasks without deterministic verification, and agents that produce irreversible real-world side effects (e.g., purchasing, email sending).

\subsection{Error Taxonomy}
\label{sec:taxonomy}

We organize benchmark defects into a structurally benchmark-agnostic taxonomy of four top-level categories (Table~\ref{tab:taxonomy-summary}), iteratively refined through audits of ScienceAgentBench and BIXBench; the full 14 subcategories are in Appendix~\ref{app:taxonomy}. While the top-level categories generalize to any execution-based benchmark, some subcategories may require extension for non-scientific domains (e.g., web navigation, OS-level tasks).

\begin{table}[t]
\centering
\caption{The \benchguard{} error taxonomy: four top-level categories of agent benchmark defects.}
\label{tab:taxonomy-summary}
\small
\begin{tabularx}{\textwidth}{@{}l c X@{}}
\toprule
\textbf{Category} & \textbf{Sub.} & \textbf{Description} \\
\midrule
\cat{GT} (Ground Truth) & 3 & Errors in the reference solution: wrong logic, wrong data handling, or format mismatches \\
\cat{EVAL} (Evaluation) & 5 & Flaws in evaluation scripts: judge bias, spec--eval mismatches, incomplete coverage, wrong tolerances, unhandled non-determinism \\
\cat{INST} (Instruction) & 3 & Defects in task specifications: underspecified requirements, internal contradictions, or infeasible tasks \\
\cat{ENV} (Environment) & 3 & Runtime configuration issues: missing dependencies, path errors, or resource constraints \\
\bottomrule
\end{tabularx}
\end{table}

Each finding produced by \benchguard{} is annotated with a severity level (Critical/High/Medium/Low) capturing impact on benchmark correctness, and a confidence score---the LLM's self-assessed probability that the defect is real---bucketed into three tiers: \emph{Confirmed} (0.8--1.0), \emph{Likely} (0.55--0.79), and \emph{Possible} (0.3--0.54); findings below 0.3 are suppressed. These operational labels are not calibrated probabilities and still require expert adjudication. Detailed definitions are in Appendix~\ref{app:taxonomy}.

\subsection{Definition-Level Audit}
\label{sec:definition}

The definition-level audit detects defects that are visible from the benchmark specification alone, without requiring any agent execution.
It combines LLM-driven verification protocols with deterministic static checks.

\paragraph{LLM Verification Protocols.}
The core of the definition-level audit is a single consolidated LLM call that cross-verifies all four artifact types, with an explicit deduplication phase to prevent duplicate findings when a single root cause manifests across multiple categories.

The protocol employs six-phase structured reasoning:
(1)~\textbf{Task Understanding}: parse the instruction to identify the target output specification;
(2)~\textbf{Ground Truth Correctness}: verify the gold program implements the correct algorithm, metrics, and data handling;
(3)~\textbf{Evaluation Logic}: verify the evaluation script measures what the specification asks, checking for judge bias, spec--eval mismatches, and incomplete coverage;
(4)~\textbf{Task Specification}: check for underspecified requirements, internal contradictions, and infeasible tasks;
(5)~\textbf{Environment}: flag runtime issues such as path mismatches and missing dependencies; and
(6)~\textbf{Consolidation}: deduplicate via the \emph{one-fix test} (if one fix would resolve multiple findings, report the root cause once), verify each finding is atomic (the \emph{split test}), filter runtime-specific artifacts, and emit the final structured report.
The system prompt includes the full taxonomy as a reference, and the user prompt presents all available task artifacts.
A key design choice is the \emph{categorization priority}: when an issue could be classified as either an instruction or ground-truth defect, the protocol defaults to the instruction-level category. Ground-truth programs are typically derived from existing codebases and are inherently more complete; the instruction is where implicit assumptions most often go undocumented---a direct consequence of solution fixation (Section~\ref{sec:intro}).

\paragraph{Execution-Level Audit.}
\label{sec:execution}
When agent traces are available, \benchguard{} runs an \emph{execution-level audit} that appends the agent's generated program, evaluation result, and evaluation log to the audit context.
Agent programs exercise code paths and edge cases invisible to static review---for instance, revealing that a tolerance is too strict only when a correct-but-differently-ordered output is scored.
On SAB, this improves recall (e.g., Opus~4.6 from 83.3\% to 91.7\%; parenthetical values in Table~\ref{tab:sab-summary}); on BIXBench, execution-level results are mixed across models and reported in Appendix~\ref{app:bix-execution}.

\section{Experimental Setup}
\label{sec:setup}

\paragraph{Benchmarks.}
We evaluate \benchguard{} on two agent benchmarks selected to cover complementary evaluation paradigms and quality-assurance histories:
\begin{itemize}[leftmargin=*,itemsep=2pt]
  \item \textbf{ScienceAgentBench (SAB)}~\citep{chen2024scienceagentbench}: 102 scientific data-analysis tasks spanning computational biology, geoscience, psychology, and materials science. Each task is evaluated by a \emph{deterministic Python script} that compares agent output to the gold program's output using domain-specific metrics. The benchmark was \emph{validated through multiple rounds of review by its creators and nine subject-matter experts}. We measure coverage of the 12 defects discovered through our auditing process and subsequently confirmed by the benchmark authors; this is a confirmed-discovery set, not an independent recall oracle.
  \item \textbf{BIXBench}~\citep{mitchener2025bixbench}: 205 bioinformatics tasks evaluated via a combination of \emph{LLM judges}, exact string matching, and numerical range verification. The benchmark underwent \emph{independent review prior to publication}, providing an external expert-revision set for estimating recall. We audit the Verified-50 subset~\citep{bixbenchverified50}---50 tasks that were independently re-reviewed by domain experts, of which 17 received revised question text, corrected ground-truth answers, or both. A \benchguard{} finding is \emph{aligned} if it identifies the same root cause as the human revision, \emph{partial} if it flags the correct functional area without pinpointing the exact issue, and \emph{unrelated} otherwise. We atomically decompose the 17 expert revisions into 24 independent issues (Appendix~\ref{app:atomic}) for finer-grained alignment matching.
\end{itemize}

\paragraph{LLM Backends.}
To assess robustness across model families, we run the definition-level audit with five frontier LLM backends, all accessed through the \code{litellm} library:
(1)~Gemini~3.0 Flash (Google, via Vertex~AI),
(2)~Gemini~3.1 Pro (Google, via Vertex~AI),
(3)~GPT-5.4 (OpenAI),
(4)~Claude Opus~4.6 (Anthropic),
(5)~Claude Sonnet~4.6 (Anthropic).
Using multiple models allows us to assess cross-model agreement and ensemble strategies.
\paragraph{Configuration.}
All audits use definition mode with deterministic checks enabled, temperature~$= 0.0$ for reproducibility (temperature~$= 1.0$ for Gemini models, which do not support~$0.0$; these models were run once, so results may vary across runs), and a maximum of 4{,}096 output tokens per LLM call.
Each task is audited independently in a single LLM call via the \code{DefinitionProtocol}.

\paragraph{Evaluation Metrics.}
We evaluate \benchguard{} along two axes:
\begin{itemize}[leftmargin=*,itemsep=2pt]
  \item \textbf{Known-issue recovery}: the fraction of known issues that \benchguard{} identifies. An LLM alignment judge (Appendix~\ref{app:judge-prompt}) classifies each (finding, known issue) pair as \emph{aligned}, \emph{partial}, or \emph{unrelated}. For SAB, this measures coverage of the 12 author-confirmed discoveries; for BIXBench, it measures recall against 24 independently produced atomic expert-revision issues. We report exact alignment (A) and partial-or-better alignment (A+P).
  \item \textbf{Finding alignment}: the fraction of reported findings on known-issue tasks that align exactly or partially with the known set. This is an agreement measure, not true precision: neither confirmed discoveries nor expert revisions necessarily enumerate every latent defect.
\end{itemize}

\paragraph{Validation Protocol.}
For SAB, the original benchmark authors independently confirmed the 12 reported defects. For BIXBench, the expert revisions were produced independently of \benchguard{}; the alignment judge performs pairwise root-cause matching but does not adjudicate whether an extra finding is a defect. We manually checked every exact-aligned pair used in the headline BIXBench recall against benchmark-owned artifacts and found no incorrect exact match (Appendix~\ref{app:additional-validation}).

\section{Results and Analysis}
\label{sec:results}

\subsection{Exposing Measurement Noise in ScienceAgentBench}
\label{sec:results-sab}

We evaluate \benchguard{}'s ability to recover the 12 author-confirmed defects in ScienceAgentBench.
These are not marginal quality issues or stylistic preferences: they include fatal specification errors that render tasks completely unsolvable, metric mismatches that silently corrupt scores, and evaluation logic that rejects correct solutions.
Coverage is measured against these 12 defects, which were discovered during auditing and subsequently confirmed by the benchmark authors; additional latent defects may exist beyond this confirmed-discovery set.
Table~\ref{tab:sab-summary} summarizes per-model recovery and finding alignment; the detailed per-defect breakdown is in Appendix~\ref{app:per-bug-sab}, which also shows that models exhibit complementary strengths across error categories.

\begin{table}[t]
\centering
\caption{Recovery and finding alignment on ScienceAgentBench (12 author-confirmed discoveries, 102 tasks).
Alignment is computed over findings within tasks containing confirmed defects, counting aligned and partial matches.
Find.\ = number of findings on defective tasks.
Cost is per-model total in USD (definition-only\,/\,+\,agent).}
\label{tab:sab-summary}
\footnotesize
\setlength{\tabcolsep}{3.2pt}
\renewcommand{\arraystretch}{1.05}
\begin{tabular}{@{}l cccc cccc rr@{}}
\toprule
& \multicolumn{4}{c}{\textbf{Definition-Only}} & \multicolumn{4}{c}{\textbf{+ Agent Programs}} & \multicolumn{2}{c}{\textbf{Cost (\$)}} \\
\cmidrule(lr){2-5} \cmidrule(lr){6-9} \cmidrule(lr){10-11}
\textbf{Model} & \textbf{Rec\textsubscript{A}} & \textbf{Rec\textsubscript{A+P}} & \textbf{Align.} & \textbf{Find.} & \textbf{Rec\textsubscript{A}} & \textbf{Rec\textsubscript{A+P}} & \textbf{Align.} & \textbf{Find.} & \textbf{Def} & \textbf{+Ag} \\
\midrule
Gemini~3.0 Flash  & 58.3 & 75.0 & 41.2 & 34 & 66.7 & 91.7 & 50.0 & 46 & 0.84 & 0.95 \\
Gemini~3.1 Pro    & 58.3 & 75.0 & \textbf{57.9} & 19 & 58.3 & 83.3 & \textbf{70.4} & 27 & 3.67 & 5.82 \\
GPT-5.4           & 58.3 & 75.0 & 47.1 & 34 & 83.3 & 91.7 & 60.0 & 45 & 2.71 & 3.10 \\
Opus~4.6          & \textbf{83.3} & \textbf{91.7} & 58.1 & 31 & \textbf{91.7} & \textbf{100} & 67.6 & 37 & 9.50 & 10.94 \\
Sonnet~4.6        & 58.3 & \textbf{91.7} & 51.9 & 27 & 83.3 & 91.7 & 59.0 & 39 & 6.00 & 6.76 \\
\midrule
\textbf{Ensemble} & 91.7 & \textbf{100} & --- & --- & \textbf{100} & \textbf{100} & --- & --- & 22.72 & 27.57 \\
\bottomrule
\end{tabular}
\end{table}

\paragraph{Finding Alignment in Context.}
The flagged-task alignment in Table~\ref{tab:sab-summary} conditions on tasks that contain confirmed defects; it measures how well practitioners can act on findings once a defective task is identified.
For a human-in-the-loop auditing tool, recovery is the primary objective: missing a genuine defect risks corrupting leaderboard conclusions, whereas non-aligned findings can be triaged using cited artifact evidence (Appendix~\ref{app:additional-validation}).

\subsection{Validating the Boundary of AI Auditing on BIXBench}
\label{sec:results-bix}

BIXBench presents a distinct evaluation paradigm: tasks are evaluated by a mix of LLM judges and deterministic verifiers (string matching, range checks), introducing new classes of potential defects---particularly in LLM-judge rubrics.
We validate \benchguard{} against the BIXBench Verified-50 subset, where 17 of 50 tasks were revised by independent domain experts.
All BIXBench results use definition-level auditing; execution-level results are reported in Appendix~\ref{app:bix-execution}.
We decompose these revisions into 24 atomic issues to enable fine-grained recall and finding-alignment measurement.

\paragraph{Overall Findings.}
Across the 50-task subset, each of the five auditor models surfaces between 43 and 114 findings (Table~\ref{tab:bix-findings}), with \subcat{INST-INCOMPLETE} consistently dominating across all models and \subcat{GT-LOGIC} and \subcat{EVAL-COVERAGE} as frequent secondary categories. The most frequent issue is underspecified instructions and evaluation rubrics anchored to a single reference implementation, penalizing valid alternative approaches.
Cases~3--4 illustrate representative defects where gold programs embed undocumented methodological choices absent from the instruction.

\begin{tcbraster}[raster columns=2, raster equal height=rows,
  raster column skip=2.5mm, raster row skip=3mm]

\begin{casebox}{Case 3: Silent Data Dropping}{\subcat{INST}}
\textbf{Instruction:}\\
\textit{``...Pearson correlation between gene length and mean expression for protein-coding genes...''}\\[0.3em]
\textbf{Gold Program:}\\
\texttt{df\_filtered = df[df.sum(axis=1) \hlcode{>= 10}]}\\[0.3em]
{\color{caseFrame}\textbf{Defect:}} Undocumented expression filter absent from the instruction.
\end{casebox}
\begin{casebox}{Case 4: Undocumented Covariates}{\subcat{INST}}
\textbf{Instruction:}\\
\textit{``...differential expression analysis... comparing \hltext{disease (ASXL1) vs control}...''}\\[0.3em]
\textbf{Gold Program:}\\
\texttt{design=\char`\~{}\hlcode{sex} + condition}\\[0.3em]
{\color{caseFrame}\textbf{Defect:}} Gold program includes \texttt{sex} covariate not mentioned in the instruction.
\end{casebox}

\end{tcbraster}

\paragraph{Alignment with Human Expert Revisions.}
Using the 17 expert-revised tasks as an independent revision set, an LLM judge classifies each \benchguard{} finding against each atomic issue as \emph{aligned} (identifies the same root cause), \emph{partial} (flags the correct area without pinpointing the exact issue), or \emph{unrelated}.
Table~\ref{tab:bix-alignment} reports per-model alignment and audit cost.
The five-model union achieves 83.3\% exact alignment (20/24 issues detected by at least one model) and 95.8\% recall when including partial matches (23/24), while the best single model (Opus~4.6) exactly aligns on 13 of 24 issues (54.2\%).
Only one issue (an ambiguous ``frequency ratio'' definition in \code{bix-54-q7}) eludes all models entirely; the remaining partially matched issues involve domain-specific conventions rather than structural cross-referencing of artifacts.
We explore the practical collaboration boundary between human and automated auditing in Cases~5--6 below.

\begin{table}[t]
\centering
\caption{Alignment with human expert revisions and audit cost on BIXBench Verified-50. Metrics are measured against 24 atomic issues from 17 expert-revised tasks; finding alignment is measured on flagged tasks (not all 50). Ensemble findings are pooled without cross-model deduplication. Cost is for the full 50-task audit.}
\label{tab:bix-alignment}
\small
\begin{tabular}{@{}l cc cc r r@{}}
\toprule
\textbf{Model} & \textbf{Recall\textsubscript{A}} & \textbf{Recall\textsubscript{A+P}} & \textbf{Align.\textsubscript{A}} & \textbf{Align.\textsubscript{A+P}} & \textbf{Cost (\$)} & \textbf{Find.} \\
\midrule
Gemini~3.0 Flash  & 45.8 & \textbf{95.8} & 33.3 & \textbf{83.3} & 0.53 & 114 \\
Gemini~3.1 Pro    & 37.5 & 58.3 & \textbf{47.1} & 76.5 & 2.31 &  43 \\
GPT-5.4           & 50.0 & 87.5 & 23.3 & 55.8 & 1.92 & 102 \\
Opus~4.6          & \textbf{54.2} & 79.2 & 38.7 & 67.7 & 5.98 &  66 \\
Sonnet~4.6        & 33.3 & 58.3 & 23.3 & 60.0 & 3.64 &  58 \\
\midrule
\textbf{Ensemble (any)} & \textbf{83.3} & \textbf{95.8} & --- & --- & 14.38 & 383 \\
\bottomrule
\end{tabular}
\end{table}

\begin{tcbraster}[raster columns=2, raster equal height=rows,
  raster column skip=2.5mm, raster row skip=3mm]

\begin{boundarybox}{Case 5: Format Over-Prescription}{\subcat{INST}}
\textbf{Original Question:}\\
\textit{``What is the difference between median treeness values for fungi versus animals?''}\\[0.3em]
\textbf{Expert Added:}\\
\textit{``Report the difference as a } \bhlcode{decimal proportion (not percentage)}\textit{.''}\\[0.3em]
{\color{boundFrame}\textbf{Partial match.}} The LLM judge accepts both formats; the added constraint is arguably unnecessary.
\end{boundarybox}
\begin{boundarybox}{Case 6: Undocumented Sample Removal}{\subcat{INST}}
\textbf{Instruction:}\\
\textit{``...\bhltext{all} significantly differentially expressed genes...''}\\[0.3em]
\textbf{Gold Program:}\\
\texttt{drop samples \bhlcode{KL3, WL3}}\\[0.3em]
{\color{boundFrame}\textbf{Caught by all 5 models, missed by experts.}} Gold silently removes two samples not mentioned in the instruction.
\end{boundarybox}

\end{tcbraster}

\paragraph{The Collaboration Boundary.}
Cases~5--6 illustrate where human and automated auditing have complementary blind spots.
In Case~5, the expert revision added a format constraint to a question whose LLM judge already accepts equivalent representations---an arguably over-prescriptive revision where \benchguard{}'s partial match reflects appropriate restraint.
In Case~6, the gold program silently drops two samples not mentioned in the instruction; all five models flag this with high severity, yet the expert review did not annotate it---likely because the gold-program author knew the removal was methodologically motivated.
The collaboration boundary runs both ways: experts can over-correct where automated auditing shows restraint, and automated auditing catches structural inconsistencies that solution fixation renders invisible to experts.
Because finding alignment is measured against expert revisions, such novel detections are counted as unaligned; this is therefore a \emph{human alignment rate}, not true precision.

\subsection{Complementary Diagnostic Profiles}
\label{sec:results-cross}

Models exhibit complementary diagnostic strengths with substantial but imperfect overlap in their detection sets (Appendix~\ref{app:ensemble}).
The five-model union achieves 83.3\% exact alignment (20/24 issues), compared to 54.2\% for the best single model, with each model contributing unique detections (detailed breakdowns in Table~\ref{tab:bix-findings}).
The entire five-model audit of 50 tasks costs under \$15 and completes in under 12 minutes, making multi-model ensembles practical.

\subsection{Additional Validation and Generality}
\label{sec:additional-validation}

The agentic skill workflow operates on benchmark-native repositories: a coding agent discovers the four artifact classes and performs any needed adaptation or conversion. On the structurally different ProgramBench cleanroom benchmark~\citep{yang2026programbench}, it audited 200 completed tasks and surfaced 114 likely findings across 45 tasks, including 26 high-severity findings across 10 tasks. Maintainers confirmed a subset during the discussion period and were triaging the remainder; with no subsequent public adjudication or corrected release, we report these as preliminary, maintainer-engaged findings rather than confirmed defects.

We also triaged every non-aligned BIXBench finding. The 272 per-model rows comprise evaluator/scoring warnings, scientific-decision sensitivity, environment/data conventions, low-priority clarifications, and 24 strong possible inconsistencies that deduplicate to five manually checked task-level clusters. A full pass took under 60 minutes. Finally, three independent alignment judges agree on 20/24 BIXBench issues exactly and 23/24 at partial-or-better, while corrected ScienceAgentBench scores shift in both directions across model settings~\citep{moussa2026d3gym}. Appendix~\ref{app:additional-validation} reports the complete analyses.

\section{Discussion and Conclusion}
\label{sec:discussion}

Our audits of two prominent, human-reviewed benchmarks show that benchmark defects are far more prevalent than commonly assumed.
\benchguard{} uncovered 12 defects in ScienceAgentBench that were later confirmed by the benchmark authors, including fatal errors that rendered tasks unsolvable, and exactly matched 83.3\% of independent expert revisions on BIXBench while surfacing strong possible artifact inconsistencies that prior human review missed.
These results suggest that even multiple rounds of careful expert validation may be insufficient to ensure benchmark integrity, and that automated auditing is a valuable complement to human review.
LLM-based auditing is not without limitations---findings require expert review, confidence and severity labels are not calibrated, and web, OS, interactive tool-use, and repository-level issue-resolution benchmarks remain future work---but at under \$15 for a five-model audit of 50 complex tasks, the cost is negligible compared to that of publishing benchmarks with undetected measurement errors.

Beyond defect detection, automated audit findings can directly inform benchmark revision, providing actionable evidence for repairing instructions, gold programs, and evaluation logic.
More importantly, auditing need not remain a post-hoc remediation step: integrating it into the benchmark construction process itself would surface implicit assumptions before they propagate into published artifacts.
This points toward a broader opportunity: if frontier models can reliably audit evaluation infrastructure, they can also serve as quality gates within automated benchmark construction pipelines---enabling the community to scale benchmark development from domain literature with built-in integrity guarantees.
Reliable evaluation at scale is ultimately a prerequisite for reliable capability improvement; ensuring that benchmarks measure what they claim to measure is foundational to the progress they are designed to track.

\section*{LLM Disclosure}
This work uses frontier LLMs as core research instruments (benchmark auditors and alignment judges), as described in Sections~\ref{sec:method}--\ref{sec:results}.
Additionally, LLMs were used to review and refine drafts of this manuscript; all scientific claims, experimental design, and analysis were conducted by the authors.

\bibliographystyle{colm2026_conference}
\bibliography{references}

\appendix

\section{Full Error Taxonomy}
\label{app:taxonomy}

Table~\ref{tab:taxonomy-full} presents the complete \benchguard{} error taxonomy with all 14 subcategories.

\begin{table}[t]
\centering
\caption{Complete \benchguard{} error taxonomy with all subcategories.}
\label{tab:taxonomy-full}
\footnotesize
\setlength{\tabcolsep}{4pt}
\renewcommand{\arraystretch}{0.97}
\begin{tabularx}{\textwidth}{@{}l p{0.29\textwidth} X@{}}
\toprule
\textbf{ID} & \textbf{Name} & \textbf{Description} \\
\midrule
\multicolumn{3}{@{}l}{\textbf{\cat{GT} (Ground Truth)}} \\
\subcat{GT-LOGIC} & Wrong logic / methodology & Gold uses incorrect algorithm, computes wrong metric, or logical opposite \\
\subcat{GT-DATA}  & Wrong data handling & Gold uses wrong files/columns, drops data, or covers only partial scope \\
\subcat{GT-FMT}   & Output format mismatch & Gold output format does not match the specification \\
\addlinespace[2pt]
\midrule
\multicolumn{3}{@{}l}{\textbf{\cat{EVAL} (Evaluation)}} \\
\subcat{EVAL-JUDGE-BIAS} & Judge bias / anchoring & Evaluator penalizes valid functionally-equivalent alternatives \\
\subcat{EVAL-MISMATCH}   & Spec--eval mismatch & Eval checks something different from the specification \\
\subcat{EVAL-COVERAGE}   & Incomplete output coverage & Eval doesn't handle all valid output formats \\
\subcat{EVAL-TOLERANCE}  & Wrong tolerance & Numeric tolerance or thresholds are incorrect \\
\subcat{EVAL-STOCHASTIC} & Unhandled non-determinism & Eval assumes deterministic outputs when it should not \\
\addlinespace[2pt]
\midrule
\multicolumn{3}{@{}l}{\textbf{\cat{INST} (Instruction)}} \\
\subcat{INST-INCOMPLETE}  & Underspecified requirements & Essential information missing, preventing a unique solution \\
\subcat{INST-CONTRADICT}  & Cross-artifact misalignment & Instructions conflict internally or with gold/eval \\
\subcat{INST-INFEASIBLE}  & Task infeasible as written & Task cannot be solved with provided information \\
\addlinespace[2pt]
\midrule
\multicolumn{3}{@{}l}{\textbf{\cat{ENV} (Environment)}} \\
\subcat{ENV-DEP}      & Missing/conflicting dependencies & Required packages unavailable or conflict \\
\subcat{ENV-PATH}     & Path configuration errors & Hardcoded paths don't match runtime environment \\
\subcat{ENV-RESOURCE} & Resource constraints & Task requires network, external services, or exceeds time limits \\
\bottomrule
\end{tabularx}
\end{table}

\paragraph{Severity and Confidence Definitions.}
\textbf{Severity} captures the impact on benchmark correctness:
\sev{Critical}---task is impossible or fundamentally broken;
\sev{High}---changes evaluation correctness for valid solutions;
\sev{Medium}---affects some valid solution strategies;
\sev{Low}---minor issue or edge case.
\textbf{Confidence} is the LLM's self-assessed probability that the defect is real, bucketed into three tiers:
\emph{Confirmed} ($0.8$--$1.0$)---verified with conclusive evidence;
\emph{Likely} ($0.55$--$0.79$)---strong evidence, high probability;
\emph{Possible} ($0.3$--$0.54$)---suspicious pattern that warrants human review.
Findings with confidence below $0.3$ are suppressed.

\clearpage
\section{Key Prompt Templates}
\label{app:prompts}

This appendix presents the key prompt templates used in \benchguard{}'s verification protocols.
The actual prompts additionally include the full error taxonomy (Table~\ref{tab:taxonomy-full}) as an inline reference and benchmark-specific calibration rules (e.g., runtime-mounted data heuristics for containerized tasks); these are omitted here for brevity.
Editorial omissions within the prompt boxes are marked with [\textit{\ldots}].
Complete prompt templates will be released with the open-source framework upon publication.

\paragraph{DefinitionProtocol System Prompt.}
The system prompt establishes the auditor role, specifies four audit areas with their subcategory checks, and defines quality-control rules governing finding granularity and calibration.

\begin{promptbox}{DefinitionProtocol --- System Prompt (condensed)}
\textbf{Role.} You are an expert benchmark auditor performing a comprehensive definition audit. Your job is to find bugs in the BENCHMARK (not in agents). You will check four areas in a single pass: ground truth correctness, evaluation logic, task specification quality, and environment issues.

\medskip
\textbf{Area 1: Ground Truth Correctness.} Verify the gold program correctly implements the task instruction.
\begin{itemize}[nosep,leftmargin=*]
  \item \texttt{GT-LOGIC}: Gold uses incorrect algorithm, computes wrong metric, or applies logical opposite.
  \item \texttt{GT-DATA}: Gold uses wrong input files or columns, drops data, or covers only partial scope.
  \item \texttt{GT-FMT}: Gold output format does not match the specification.
\end{itemize}

\textbf{Area 2: Evaluation Logic.} Verify the evaluation script correctly measures what the specification asks.
\begin{itemize}[nosep,leftmargin=*]
  \item \texttt{EVAL-JUDGE-BIAS}: LLM judge rigidly anchored to one implementation, rejecting valid alternatives.
  \item \texttt{EVAL-MISMATCH}: Eval checks something different from what the specification requests.
  \item \texttt{EVAL-COVERAGE}: Eval does not handle all valid output formats, types, or equivalent names.
  \item \texttt{EVAL-TOLERANCE}: Numerical tolerances too strict or too lenient.
  \item \texttt{EVAL-STOCHASTIC}: Eval assumes deterministic output for inherently non-deterministic computation.
\end{itemize}

\textbf{Area 3: Task Specification.} Check whether the instruction provides sufficient and consistent information.
\begin{itemize}[nosep,leftmargin=*]
  \item \texttt{INST-INCOMPLETE}: Essential information missing, preventing a unique correct solution.
  \item \texttt{INST-CONTRADICT}: Instruction conflicts with the gold program or evaluation script.
  \item \texttt{INST-INFEASIBLE}: Task cannot be solved with the provided information.
\end{itemize}

\textbf{Area 4: Environment \& Infrastructure.} Flag runtime issues in a sandboxed evaluation environment.
\begin{itemize}[nosep,leftmargin=*]
  \item \texttt{ENV-DEP}: Required packages unavailable or version conflicts.
  \item \texttt{ENV-PATH}: Hardcoded absolute paths that do not match the evaluation environment.
  \item \texttt{ENV-RESOURCE}: Requires network access, external APIs, or exceeds time/compute limits.
\end{itemize}

[\textit{Per-area guidelines and benchmark-specific calibration rules omitted.}]

\medskip
\textbf{Categorization Priority.} When an issue could be classified as either an instruction problem (\texttt{INST-*}) or a ground truth problem (\texttt{GT-*}), always prefer the instruction-level category. The root cause of most benchmark bugs is an underspecified or ambiguous instruction---the gold program simply implements one interpretation of the vague spec. Use \texttt{GT-*} only when the gold program is objectively wrong independent of the instruction (e.g., a coding bug, off-by-one error, wrong formula).

\medskip
\textbf{Deduplication Rule.} Each bug must appear exactly once under its most specific subcategory. Apply the \emph{one-fix test}: if fixing one issue would make another finding disappear, they are the same bug---report it only once.

\medskip
\textbf{Atomicity Rule.} Every finding MUST describe exactly one independently fixable root cause. Apply the \emph{split test} to every finding: ``Could concern A be fixed while concern B still remains?'' If yes, they must be separate findings.

\medskip
\textbf{Output Format.} Respond with a JSON array. Each finding includes:
\texttt{category} (GT/EVAL/INST/ENV),
\texttt{subcategory} (e.g., \texttt{GT-LOGIC}),
\texttt{severity},
\texttt{finding\_type},
\texttt{title},
\texttt{description},
\texttt{evidence} (with source file, line numbers, and code snippet),
\texttt{recommendation}, and
\texttt{confidence} (0--1).

\smallskip
\emph{Severity scale:}
Critical = task is impossible or fundamentally broken;
High = changes evaluation correctness;
Medium = affects some valid solutions;
Low = minor issue or edge case.

\smallskip
\emph{Finding type:}
\textsc{Bug} = a concrete, verifiable error---a specific line of code contradicts the specification, an incorrect algorithm is used, or logic is demonstrably broken.
\textsc{Warning} = a concern or improvement opportunity---something is underspecified or not provably wrong. Default to \textsc{Warning} when concrete scoring harm cannot be demonstrated.

\smallskip
\emph{Confidence calibration:}
0.8--1.0 (\textsc{Confirmed}) = a specific line contradicts a specific statement in the spec;
0.55--0.79 (\textsc{Likely}) = clear discrepancy requiring cross-section reasoning or domain inference;
0.3--0.54 (\textsc{Possible}) = suspicious pattern that requires domain expert verification.
Below 0.3: do not report.

\smallskip
[\textit{Full JSON schema, inline taxonomy reference, and runtime-data calibration rules omitted for brevity.}]
\end{promptbox}

\paragraph{DefinitionProtocol User Prompt.}
The user prompt presents all task artifacts in a structured context block, followed by a six-phase chain-of-thought analysis procedure.
Phases~2--5 produce analysis only (no JSON output); Phase~6 consolidates, deduplicates, and emits the final JSON array.

\begin{promptbox}{DefinitionProtocol --- User Prompt Template}
\textbf{Context Block} [\textit{populated per task}]\textbf{:}\\[0.3em]
\texttt{Task ID:} \{task\_id\}~~\textbar{}~~\texttt{Domain:} \{domain\}~~\textbar{}~~\texttt{Expected output:} \{expected\_output\}\\[0.3em]
\texttt{Task Instruction:}~~\{task\_instruction\}\\
\texttt{Gold Program} [\textit{with line numbers}]\texttt{:}~~\{gold\_program\_source\}\\
\texttt{Evaluation Script} [\textit{with line numbers}]\texttt{:}~~\{eval\_script\_source\}\\
\texttt{Input Data Description:}~~\{input\_data\_description\}\\
\texttt{Domain Knowledge} [\textit{optional}]\texttt{:}~~\{domain\_knowledge\}\\
\texttt{Environment Information:}~~\{environment\_info\}

\medskip
\textbf{Phase 1: Understand the Task.}
Parse the instruction to identify target output, metric, and computation. Classify the runtime environment (container/capsule vs.\ standalone) to calibrate discoverability of data details.

\smallskip
\textbf{Phase 2: Ground Truth Correctness} [\textit{analysis only}]\textbf{.}
Trace the gold program line by line. Compare against instruction: correct metric? Correct files and columns? Right algorithm? Correct output format? List each concern as a separate bullet.

\smallskip
\textbf{Phase 3: Evaluation Logic} [\textit{analysis only}]\textbf{.}
Trace the evaluation script. Does it check what the spec asks? Handle all valid output formats? Use appropriate tolerances? List concerns separately.

\smallskip
\textbf{Phase 4: Task Specification} [\textit{analysis only}]\textbf{.}
For each candidate omission, apply a discoverability gate: is this a runtime-discoverable detail (filenames, column names, metadata) or a non-discoverable semantic choice (methodology, thresholds, covariates)? Report only the latter as genuine concerns.

\smallskip
\textbf{Phase 5: Environment \& Infrastructure} [\textit{analysis only}]\textbf{.}
Check for path mismatches, resource constraints, and sandbox compatibility.

\smallskip
\textbf{Phase 6: Consolidate \& Report.}
(1)~Deduplicate via the one-fix test.
(2)~Atomicity check via the split test.
(3)~Runtime-data filter: remove findings about purely discoverable details.
(4)~Output the final JSON array.
\end{promptbox}

\paragraph{Agent Program as Supporting Evidence.}
When agent trajectories are available, the audit extends to \emph{execution-level mode} by incorporating the agent's generated program as supporting evidence.
This mode corresponds to the execution-level results in Table~\ref{tab:sab-summary}: agent programs exercise code paths and edge cases invisible to static review, improving recall (e.g., Opus~4.6 from 83.3\% to 91.7\% on SAB).
The agent program is appended to the user prompt as an additional section, and Phase~1 is augmented accordingly.

\begin{promptbox}{Agent Program Evidence --- User Prompt Extension}
\textbf{Agent Program (Supporting Evidence Only)}

Use this as supporting evidence about how a concrete agent interpreted the task. Do NOT assume the agent is correct. Use the benchmark artifacts as the primary source of truth, and only use the agent program to expose hidden assumptions, ambiguity, evaluator anchoring, or other benchmark issues that become clearer when compared against a real candidate solution.

\medskip
\texttt{[Agent's generated program source code]}

\medskip
\textbf{Phase 1 augmentation:} ``If an agent program is provided below, treat it as supporting evidence about one plausible interpretation of the task, not as ground truth.''
\end{promptbox}

\section{Benchmark Ingestion Details}
\label{app:ingestion}

The reference Python pipeline uses a standardized directory format compatible with Harbor~\citep{harbor2026}. The agentic skill workflow is not tied to this layout: a coding agent can inspect native benchmark files, identify the four artifact classes, and perform the conversion or adaptation needed to emit the same finding schema.
Each subdirectory under the benchmark root is treated as a task if and only if it contains a \code{task.toml} file.

\begin{lstlisting}[language={}, numbers=none, caption={Benchmark directory layout.}, label={lst:layout}]
benchmark_root/
|- benchguard_hints.yaml        # optional: benchmark-level review hints
|- task_id/
|  |- task.toml                 # required: metadata + config
|  |- instruction.md            # required: natural-language task spec
|  |- tests/                    # required: evaluation logic
|  |  |- test.sh
|  |  `- *.py, *.sh, *.json, ...
|  |- solution/                 # recommended: reference solution
|  |  |- solve.sh
|  |  `- *.py, *.sh, ...
|  |- environment/              # optional: Dockerfile, requirements
|  |  |- Dockerfile
|  |  `- requirements.txt
|  |- domain_knowledge.md       # optional: background context
|  `- data_description.md       # optional: dataset inventory
`- another_task_id/
   `- ...
\end{lstlisting}

The \code{task.toml} file stores benchmark metadata and runtime configuration.
All fields are optional; the only structural requirement is that the file exists.

\begin{lstlisting}[language={}, numbers=none, caption={Representative \code{task.toml} configuration.}, label={lst:toml}]
[metadata]
id = "task_001"
category = "bioinformatics"
expected_output = "output.csv"
benchmark_source = "ScienceAgentBench"

[verifier]
method = "script"

[environment]
runtime = "python3.10"
cpus = 2
memory = "4G"

[agent]
timeout_sec = 1800.0
\end{lstlisting}

\paragraph{Tiered Input.}
The framework supports three input tiers of increasing diagnostic power:
\begin{itemize}[leftmargin=*,itemsep=1pt]
  \item \textbf{Minimal} (instruction + tests): enables \cat{EVAL}, \cat{INST}, and \cat{ENV} checks.
  \item \textbf{Definition-level} (+ gold program): enables full cross-artifact auditing.
  \item \textbf{Execution-level} (+ agent results): enables trajectory-informed verification.
\end{itemize}

\paragraph{File Loading.}
When loading a task, \benchguard{} concatenates text files in a deterministic order: \code{test.sh}, \code{solve.sh}, and \code{Dockerfile} are loaded first in their respective directories, followed by remaining text files in alphabetical order; binary files are ignored.
A root-level \code{benchguard\_hints.yaml} file can optionally inject benchmark-specific review guidance into the audit prompts.
Converter scripts transform existing benchmark formats into this standard layout. In the skill workflow, the coding agent performs this adaptation, so benchmark authors need not manually reformat their repositories.

\section{Per-Task Detailed Findings}
\label{app:per-task}

Complete per-task findings for all 102 ScienceAgentBench tasks and 50 BIXBench Verified-50 tasks---including finding descriptions, evidence, and confidence scores---will be released alongside the open-source framework upon publication.

\section{Per-Defect Recall on ScienceAgentBench}
\label{app:per-bug-sab}

Table~\ref{tab:sab-per-bug} lists each of the 12 author-confirmed defects in ScienceAgentBench and indicates which models detected them.

\begin{table}[t]
\centering
\caption{Per-defect recall on ScienceAgentBench's 12 confirmed defects (execution-level: with agent programs as additional input). \cmark{} = aligned (exact match), $\sim$ = partial match, \xmark{} = missed. Recall is computed on exact matches only. Definition-only recall is reported in aggregate in Table~\ref{tab:sab-summary}.}
\label{tab:sab-per-bug}
\small
\begin{tabular}{@{}r l l ccccc@{}}
\toprule
\textbf{Task} & \textbf{Category} & \textbf{Description} & \rotatebox{70}{\footnotesize\textbf{Flash}} & \rotatebox{70}{\footnotesize\textbf{Pro}} & \rotatebox{70}{\footnotesize\textbf{GPT-5.4}} & \rotatebox{70}{\footnotesize\textbf{Opus~4.6}} & \rotatebox{70}{\footnotesize\textbf{Sonnet~4.6}} \\
\midrule
9   & \cat{GT}   & Pearson $r$ vs.\ $R^2$ metric      & \cmark & \cmark & \cmark & \cmark & \cmark \\
12  & \cat{EVAL} & Output format (SMILES vs.\ names)   & \cmark & \cmark & \cmark & \cmark & \cmark \\
21  & \cat{GT}   & Wrong deforestation rate             & \cmark & \cmark & \cmark & \cmark & \cmark \\
26  & \cat{INST} & Singular vs.\ plural naming          & $\sim$ & $\sim$ & \cmark & \cmark & \cmark \\
29  & \cat{INST} & Wrong input file (Critical)          & \cmark & \cmark & \cmark & \cmark & \cmark \\
31  & \cat{INST} & Unspecified analysis method           & \cmark & $\sim$ & $\sim$ & \cmark & \cmark \\
32  & \cat{INST} & Wrong analysis grouping               & $\sim$ & \xmark & \cmark & \cmark & \cmark \\
34  & \cat{INST} & Infeasible requirement                & \cmark & \cmark & \cmark & \cmark & \cmark \\
35  & \cat{INST} & Unspecified output format             & \cmark & \cmark & \cmark & \cmark & \cmark \\
67  & \cat{INST} & Wrong output save path                & \cmark & $\sim$ & \cmark & \cmark & \cmark \\
78  & \cat{GT}   & Test data contamination               & \xmark & \xmark & \xmark & \cmark & \xmark \\
92  & \cat{GT}   & Column dimension mismatch             & $\sim$ & \cmark & \cmark & $\sim$ & $\sim$ \\
\bottomrule
\end{tabular}
\end{table}

\section{Detailed Finding Distribution on BIXBench}
\label{app:bix-findings}

Table~\ref{tab:bix-findings} provides the full subcategory-level breakdown of \benchguard{} findings on BIXBench Verified-50.

\begin{table}[t]
\centering
\caption{Distribution of \benchguard{} findings on BIXBench Verified-50 (50 tasks) by subcategory and auditor model.}
\label{tab:bix-findings}
\small
\begin{tabular}{@{}l rrrrr@{}}
\toprule
\textbf{Subcategory} & \textbf{Flash} & \textbf{Pro} & \textbf{GPT-5.4} & \textbf{Opus} & \textbf{Sonnet} \\
\midrule
\subcat{GT-LOGIC}          & 21 &  3 & 19 &  3 &  7 \\
\subcat{GT-DATA}           &  1 &  0 &  2 &  0 &  0 \\
\subcat{EVAL-COVERAGE}     &  0 &  2 & 23 &  3 &  8 \\
\subcat{EVAL-MISMATCH}     &  9 &  2 &  6 &  0 &  1 \\
\subcat{EVAL-JUDGE-BIAS}   &  0 &  0 &  7 &  0 &  0 \\
\subcat{EVAL-TOLERANCE}    &  5 &  1 &  1 &  3 &  0 \\
\subcat{EVAL-STOCHASTIC}   &  0 &  0 &  6 & 20 & 10 \\
\subcat{INST-INCOMPLETE}   & 78 & 33 & 33 & 34 & 31 \\
\subcat{INST-CONTRADICT}   &  0 &  2 &  5 &  3 &  1 \\
\midrule
\textbf{Total}             & 114 & 43 & 102 & 66 & 58 \\
\bottomrule
\end{tabular}
\end{table}

\section{Ensemble Analysis and Model-Specific Patterns}
\label{app:ensemble}

\paragraph{Agreement.}
Models show substantial but imperfect overlap in their detection sets: eleven of 24 issues achieve strong consensus ($\geq$3 models exactly aligned), including ambiguous filtering criteria (\code{bix-22-q4}), multi-issue tasks (\code{bix-31-q2}), and missing covariates (\code{bix-49-q4}).

\paragraph{Ensemble Strategies.}
The union of all five models achieves 83.3\% exact alignment (20/24), compared to 54.2\% for the best single model (Opus~4.6).
Majority vote ($\geq$3 models aligned) covers 11/24 issues (45.8\%) with fewer findings to triage; relaxing to $\geq$2 models reaches 62.5\%.
When including partial matches, $\geq$3 models detect 87.5\% of issues (21/24), and $\geq$2 models reach 95.8\% (23/24).

\paragraph{Model-Specific Patterns.}
All models identify \subcat{INST-INCOMPLETE} as the dominant subcategory (31--78 findings), but secondary diagnostic profiles diverge: Gemini~3.0 Flash and GPT-5.4 surface the most \subcat{GT-LOGIC} errors (21 and 19, respectively), Claude models produce the most \subcat{EVAL-STOCHASTIC} findings (20 and 10), and GPT-5.4 is the only model detecting \subcat{EVAL-JUDGE-BIAS} (7 findings).
GPT-5.4 also produces the most \subcat{EVAL-COVERAGE} findings (23 vs.\ 0--8 for other models), suggesting greater sensitivity to evaluation infrastructure issues.

\paragraph{LLM-Judge-Specific Failure Modes.}
A finding unique to BIXBench is the prevalence of \subcat{EVAL-JUDGE-BIAS} errors---cases where the LLM judge is anchored to a single implementation strategy.
Notably, only GPT-5.4 detects these (7 findings); no other model flags any, suggesting sensitivity to judge-specific failure modes varies sharply across model families.
This failure mode is absent in script-evaluated benchmarks like SAB, where evaluation logic is deterministic.

\paragraph{Cost Breakdown.}
Gemini~3.0 Flash audits all 50 tasks for \$0.53 total (\$0.011/task) with the highest partial-match recall (95.8\%), while even the most expensive single model (Opus~4.6, \$5.98) costs less than \$0.12 per task.

\section{Atomic Issue Decomposition for BIXBench}
\label{app:atomic}

To enable fine-grained recall and finding-alignment measurement, we decompose the 17 expert-revised questions in the Verified-50 subset into 24 atomic issues.
Each atomic issue represents a single, independently fixable concern identified during expert revision, following the \emph{split test}: if concern~A can be fixed while concern~B remains, they are separate issues.
Table~\ref{tab:atomic-summary} summarizes the category distribution, and Table~\ref{tab:atomic-issues} lists all 24 issues.

\begin{table}[t]
\centering
\caption{Category distribution of 24 atomic issues in BIXBench Verified-50.}
\label{tab:atomic-summary}
\small
\begin{tabular}{@{}l l r@{}}
\toprule
\textbf{Category} & \textbf{Subcategory} & \textbf{Count} \\
\midrule
\cat{INST} & \subcat{INST-INCOMPLETE} & 21 \\
\cat{GT}   & \subcat{GT-LOGIC}        &  3 \\
\midrule
\multicolumn{2}{@{}l}{\textbf{Total}} & \textbf{24} \\
\bottomrule
\end{tabular}
\end{table}

Instruction-level issues account for 87.5\% (21/24) of all atomic issues, spanning underspecified requirements, ambiguous definitions, and missing constraints.
Two of the three \subcat{GT-LOGIC} issues co-occur with instruction-level issues in multi-issue tasks (e.g., \code{bix-31-q2} and \code{bix-43-q2}), where an underspecified instruction led the gold program to implement one particular interpretation; the third (\code{bix-31-q2}, FAM138A categorization) reflects a factual error in the gold program.

Twelve tasks contain a single issue, three tasks contain two issues, and two tasks (\code{bix-31-q2}, \code{bix-54-q7}) contain three independently fixable issues each.
In the ``Change'' column, \textbf{Q}~= question revised, \textbf{A}~= ideal answer revised, \textbf{B}~= both revised.

\begin{table}[t]
\centering
\caption{All 24 atomic issues from expert revision of BIXBench Verified-50.}
\label{tab:atomic-issues}
\small
\begin{tabularx}{\textwidth}{@{}l l l X@{}}
\toprule
\textbf{Task} & \textbf{Category} & \textbf{Change} & \textbf{Description} \\
\midrule
bix-6-q4   & \subcat{INST-INCOMPLETE}  & Q & Ambiguous correlation metric (expression, fold change, or P-values) \\
bix-11-q1  & \subcat{INST-INCOMPLETE}  & Q & Unspecified numerical format (decimal vs.\ percentage) \\
bix-14-q1  & \subcat{INST-INCOMPLETE} & B & Variant scope unspecified; should restrict to coding variants \\
bix-20-q3  & \subcat{INST-INCOMPLETE} & Q & Classification source (ClinVar) not specified \\
bix-20-q3  & \subcat{INST-INCOMPLETE}  & Q & ``Benign'' ambiguous w.r.t.\ inclusion of ``Likely Benign'' \\
bix-22-q4  & \subcat{INST-INCOMPLETE}  & Q & Ambiguous gene filtering (all vs.\ expressed protein-coding genes) \\
bix-26-q5  & \subcat{INST-INCOMPLETE}  & Q & Conflated gene-level and pathway-level significance thresholds \\
bix-27-q5  & \subcat{INST-INCOMPLETE} & B & PCA component count unspecified; gold uses 100 \\
bix-28-q3  & \subcat{INST-INCOMPLETE} & Q & Tool for long branch score calculation unspecified (PhyKIT) \\
bix-31-q2  & \subcat{INST-INCOMPLETE} & B & Batch correction strategy and software (pydeseq2) unspecified \\
bix-31-q2  & \subcat{GT-LOGIC}        & B & FAM138A incorrectly categorized as protein-coding (is lncRNA) \\
bix-31-q2  & \subcat{GT-LOGIC}        & B & Ideal answer wrong due to missing batch covariate and shrinkage method \\
bix-32-q2  & \subcat{INST-INCOMPLETE}  & Q & ``KEGG enrichment'' ambiguous between GSEA and ORA methods \\
bix-43-q2  & \subcat{INST-INCOMPLETE} & B & DESeq2 configuration details missing (treatment groups, pre-filtering) \\
bix-43-q2  & \subcat{GT-LOGIC}        & B & Threshold operators inconsistent (p vs.\ padj, strict vs.\ non-strict) \\
bix-43-q4  & \subcat{INST-INCOMPLETE}  & Q & ``Proportion of DEGs'' ambiguous (DEGs in pathway vs.\ pathway size) \\
bix-49-q4  & \subcat{INST-INCOMPLETE} & Q & Missing sex covariate requirement in DE analysis \\
bix-52-q2  & \subcat{INST-INCOMPLETE}  & Q & ``Average density'' ambiguous (global vs.\ mean of per-chromosome) \\
bix-52-q2  & \subcat{INST-INCOMPLETE} & Q & No filter for chromosomes with zero CpGs \\
bix-52-q7  & \subcat{INST-INCOMPLETE}  & Q & ``Sites'' ambiguous (unique genomic locations vs.\ data rows) \\
bix-54-q7  & \subcat{INST-INCOMPLETE} & Q & Spline degrees of freedom (df=4) unspecified \\
bix-54-q7  & \subcat{INST-INCOMPLETE}  & Q & ``Frequency ratio'' ambiguous; clarified as proportion of strain 287 \\
bix-54-q7  & \subcat{INST-INCOMPLETE} & Q & Software environment (R) unspecified for model fitting \\
bix-61-q2  & \subcat{INST-INCOMPLETE}  & Q & Coverage depth scope ambiguous (entire genome vs.\ covered regions) \\
\bottomrule
\end{tabularx}
\end{table}

\section{Execution-Level Audit on BIXBench}
\label{app:bix-execution}

We also evaluated the execution-level audit on BIXBench by appending agent-generated summaries to the audit context.
Unlike SAB---where full agent programs provide rich diagnostic signal---the execution-level results on BIXBench are mixed: some models show modest recall improvements while others see no gain or slight degradation.
This suggests that the value of execution-level auditing depends on the richness and informativeness of the available agent traces.
The main BIXBench results (Tables~\ref{tab:bix-alignment} and~\ref{tab:bix-findings}) use definition-level auditing only.

Table~\ref{tab:bix-execution-summary} compares definition-only auditing against the version that appends agent-generated execution summaries.
The effect is uneven across models.
Gemini~3.0 Flash benefits the most, improving exact recall from 45.8\% to 66.7\% while reducing total findings from 114 to 105.
Sonnet~4.6 improves exact recall but loses partial coverage; GPT-5.4 is largely unchanged in recall but has higher finding alignment; Opus~4.6 and Gemini~3.1 Pro regress in exact recall.
At the ensemble level, the five-model union improves only marginally, from 83.3\% to 87.5\% exact recall, while Recall\textsubscript{A+P} remains 95.8\%.

The only newly exact-matched issue is \code{bix-32-q2}, where the ambiguous ``KEGG enrichment'' instruction becomes easier to ground when the execution summary exposes the concrete analysis path taken by the agent.
The same issue remains completely missed by all models under both settings: \code{bix-54-q7}'s ambiguous ``frequency ratio'' definition.
Overall, compact execution summaries provide weaker incremental signal on BIXBench than full agent programs do on ScienceAgentBench, so we treat definition-level auditing as the primary result.

\begin{table}[H]
\centering
\caption{Definition-only vs.\ execution-summary auditing on BIXBench Verified-50.
Finding alignment is measured on findings within the 17 expert-revised tasks.
Find.\ = total findings across the 50-task audit.
Cost is per-model total in USD (definition-only\,/\,+\,execution summary).}
\label{tab:bix-execution-summary}
\footnotesize
\setlength{\tabcolsep}{3.2pt}
\renewcommand{\arraystretch}{1.05}
\begin{tabular}{@{}l cccc cccc rr@{}}
\toprule
\multicolumn{1}{c}{} & \multicolumn{4}{c}{\textbf{Definition-Only}} & \multicolumn{4}{c}{\textbf{+ Execution Summaries}} & \multicolumn{2}{c}{\textbf{Cost (\$)}} \\
\cmidrule(lr){2-5} \cmidrule(lr){6-9} \cmidrule(lr){10-11}
\textbf{Model} & \textbf{Rec\textsubscript{A}} & \textbf{Rec\textsubscript{A+P}} & \textbf{Al.\textsubscript{A}} & \textbf{Al.\textsubscript{A+P}} & \textbf{Rec\textsubscript{A}} & \textbf{Rec\textsubscript{A+P}} & \textbf{Al.\textsubscript{A}} & \textbf{Al.\textsubscript{A+P}} & \textbf{Def} & \textbf{+Exec} \\
\midrule
Gemini~3.0 Flash & 45.8 & \textbf{95.8} & 33.3 & \textbf{83.3} & 66.7 & \textbf{95.8} & 36.6 & 80.5 & 0.53 & 0.32 \\
Gemini~3.1 Pro   & 37.5 & 58.3 & \textbf{47.1} & 76.5 & 33.3 & 62.5 & \textbf{47.1} & \textbf{100.0} & 2.31 & 2.31 \\
GPT-5.4          & 50.0 & 87.5 & 23.3 & 55.8 & 50.0 & 87.5 & 27.3 & 68.2 & 1.92 & 1.94 \\
Opus~4.6         & \textbf{54.2} & 79.2 & 38.7 & 67.7 & 33.3 & 75.0 & 30.8 & 76.9 & 5.98 & 6.00 \\
Sonnet~4.6       & 33.3 & 58.3 & 23.3 & 60.0 & 41.7 & 50.0 & 32.3 & 45.2 & 3.64 & 3.66 \\
\midrule
\textbf{Ensemble} & 83.3 & \textbf{95.8} & --- & --- & \textbf{87.5} & \textbf{95.8} & --- & --- & 14.38 & 14.23 \\
\bottomrule
\end{tabular}
\end{table}

\section{Additional Validation Analyses}
\label{app:additional-validation}

\subsection{Agentic Skill and ProgramBench Extension}

The agentic skill applies the same definition-level methodology in Claude Code and Codex while discovering or adapting benchmark artifacts in place. Table~\ref{tab:skill-results} reports exact / partial-or-better recovery on the two paper benchmarks. The two stacks peak on different benchmarks, indicating that the skill is useful with either backend without requiring a five-model ensemble.

\begin{table}[H]
\centering
\caption{Definition-only skill results. SAB uses the 12 author-confirmed discovery set; BIXBench uses 24 independently produced atomic expert-revision issues.}
\label{tab:skill-results}
\small
\begin{tabular}{@{}l cc@{}}
\toprule
\textbf{Benchmark} & \textbf{Claude Code + skill} & \textbf{Codex + skill} \\
\midrule
ScienceAgentBench & 83.3 / 91.7 & 58.3 / 66.7 \\
BIXBench Verified-50 & 58.3 / 91.7 & 83.3 / 87.5 \\
\bottomrule
\end{tabular}
\end{table}

ProgramBench~\citep{yang2026programbench} is a non-Harbor cleanroom code-reconstruction benchmark with a substantially different native repository structure. A strict Codex skill audit covered 200 completed tasks (plus one partial metadata task) and produced 114 likely findings across 45 tasks: 26 high-severity \textsc{bug} findings across 10 tasks and 88 medium-severity warnings. The findings comprise 88 environment/configuration, 23 CLI-flag, two subcommand, and one native test-validity case. We disclosed representative findings to the maintainers, who confirmed a subset and were triaging the remainder during the discussion period. No subsequent public adjudication or corrected release was available at camera-ready time, so these counts are preliminary, maintainer-engaged findings rather than confirmed defects.

\subsection{Non-Aligned Finding Triage}

Non-aligned findings are heterogeneous: they may be suppressible warnings, legitimate implicit conventions, useful clarifications, or possible defects absent from the known set. We classify each finding into five operational buckets. \emph{Evaluator/scoring mechanics} covers generic complaints without demonstrated scoring harm; \emph{scientific decision sensitivity} disputes defensible methodological choices; \emph{environment/data convention} asks to state information already supplied by the harness; \emph{clarification} is valid but not scoring-relevant; and \emph{strong inconsistency} is a concrete, answer-changing contradiction across benchmark-owned artifacts.

\begin{table}[H]
\centering
\caption{Triage of all non-aligned BIXBench finding rows. Counts are per auditor, not unique defects. Eval.=evaluator/scoring mechanics; Sci.=scientific-decision sensitivity; Env.=environment/data convention; Clar.=low-priority clarification; Strong=strong possible artifact inconsistency.}
\label{tab:bix-triage}
\small
\setlength{\tabcolsep}{4pt}
\begin{tabular}{@{}l rrrrrr@{}}
\toprule
\textbf{Auditor} & \textbf{All} & \textbf{Eval.} & \textbf{Sci.} & \textbf{Env.} & \textbf{Clar.} & \textbf{Strong} \\
\midrule
Gemini~3.0 Flash & 79 & 18 & 22 & 23 & 12 & 4 \\
Gemini~3.1 Pro   & 30 &  6 &  6 &  7 &  6 & 5 \\
GPT-5.4          & 78 & 46 &  8 &  7 & 12 & 5 \\
Opus~4.6         & 45 & 22 &  7 &  4 &  7 & 5 \\
Sonnet~4.6       & 40 & 20 &  5 &  3 &  7 & 5 \\
\midrule
\textbf{Total rows} & \textbf{272} & \textbf{112} & \textbf{48} & \textbf{44} & \textbf{44} & \textbf{24} \\
\bottomrule
\end{tabular}
\end{table}

The 24 strong rows deduplicate to five manually checked task-level clusters: \code{bix-26-q3} (the prompt requests adjusted-$p$ filtering but the reference filters only by fold change), \code{bix-37-q4} (evaluator normalization removes a minus sign), \code{bix-51-q8} (an inverted numeric range), \code{bix-52-q6} (the prompt asks about Jackdaw while the gold/evaluator expect a Zebrafinch-like Chromosome~W answer), and \code{bix-53-q5} (the gold silently drops samples \code{KL3}/\code{WL3}). We treat them as strong possible inconsistencies, not confirmed defects.

\begin{table}[H]
\centering
\caption{Non-aligned rows on the 12 author-confirmed SAB tasks in the execution-level setting. Counts are per auditor, not unique defects.}
\label{tab:sab-triage}
\small
\setlength{\tabcolsep}{4pt}
\begin{tabular}{@{}l rrrrrr@{}}
\toprule
\textbf{Auditor} & \textbf{All} & \textbf{Eval.} & \textbf{Sci.} & \textbf{Env.} & \textbf{Clar.} & \textbf{Strong} \\
\midrule
Gemini~3.0 Flash & 23 & 11 & 6 & 3 & 0 & 3 \\
Gemini~3.1 Pro   &  8 &  2 & 3 & 2 & 0 & 1 \\
GPT-5.4          & 18 & 10 & 2 & 1 & 1 & 4 \\
Opus~4.6         & 12 &  5 & 2 & 2 & 0 & 3 \\
Sonnet~4.6       & 16 &  8 & 1 & 2 & 0 & 5 \\
\bottomrule
\end{tabular}
\end{table}

The strong SAB rows concentrate on three possible additional task-level inconsistencies (tasks 29, 32, and 78) and are not counted as author-confirmed defects. Because each finding cites its artifact class and evidence, most rows are dismissible at a glance; a complete triage pass took under 60 minutes per benchmark, with effort concentrated in the small line-level subset.

\subsection{Alignment-Judge Robustness}

The alignment judge only determines whether a finding and a known issue describe the same root cause; it does not decide whether an unaligned finding is a real defect. We re-judged the same 247 BIXBench (finding, expert issue) pairs with GPT-5.4 and Claude Sonnet~4.6 in addition to Gemini~3.0 Flash. Table~\ref{tab:judge-agreement} shows substantial pair-level agreement across model families. At the issue level, all three judges agree on the best exact verdict for 20/24 issues and on partial-or-better for 23/24. Exact / partial-or-better recall is 20/24 / 23/24 for Flash, 19/24 / 24/24 for GPT-5.4, and 17/24 / 23/24 for Sonnet, showing that partial-or-better recall is stable while exact alignment is more judge-sensitive. Every exact-aligned pair used in the headline result was manually checked against benchmark-owned artifacts; we found no incorrect exact match.

\begin{table}[H]
\centering
\caption{Pair-level alignment-judge agreement on 247 BIXBench pairs. A+P collapses aligned and partial into one class; $\kappa$ is Cohen's kappa.}
\label{tab:judge-agreement}
\small
\begin{tabular}{@{}l cc cc@{}}
\toprule
\textbf{Judge pair} & \textbf{3-label agree} & \textbf{$\kappa$} & \textbf{A+P agree} & \textbf{$\kappa$} \\
\midrule
Flash / GPT-5.4 & 87.9 & 0.811 & 91.5 & 0.828 \\
Flash / Sonnet~4.6 & 79.4 & 0.673 & 84.2 & 0.690 \\
GPT-5.4 / Sonnet~4.6 & 81.4 & 0.699 & 84.6 & 0.695 \\
\bottomrule
\end{tabular}
\end{table}

\subsection{Downstream Impact and Suggested Repairs}

ScienceAgentBench-Verified, released by the ScienceAgentBench team as part of D3-Gym~\citep{moussa2026d3gym}, repairs issues checked after being flagged with \benchguard{}. Table~\ref{tab:sab-verified-impact} shows bidirectional score shifts: upward changes indicate corrections that unblock valid solutions, whereas the downward change indicates a previously over-credited setting. This is external downstream-impact evidence, but not fully independent defect discovery because the repair process began from \benchguard{} findings.

\begin{table}[H]
\centering
\caption{Reported score shifts from SAB to SAB-Verified.}
\label{tab:sab-verified-impact}
\small
\begin{tabular}{@{}l l rrr@{}}
\toprule
\textbf{Model / setting} & \textbf{Metric} & \textbf{SAB} & \textbf{Verified} & \textbf{$\Delta$} \\
\midrule
Qwen3-8B / Base & SR & 6.5 & 8.2 & +1.7 \\
Qwen3-8B / Base & SR@3 & 11.8 & 13.8 & +2.0 \\
Qwen3-14B / Base & SR & 9.1 & 10.3 & +1.2 \\
Qwen3-8B / RFT-Self & VER@3 & 43.1 & 41.2 & $-1.9$ \\
Qwen3-32B / Base & SR & 14.7 & 15.0 & +0.3 \\
\bottomrule
\end{tabular}
\end{table}

Each \benchguard{} finding includes a suggested revision. Representative repairs include changing SAB task~29's instruction to the actual packaged input filename, aligning task~12's evaluator with the requested SMILES format (or explicitly requesting drug names), and either documenting the \code{KL3}/\code{WL3} removal in \code{bix-53-q5} or retaining all samples. Because such changes alter evaluation semantics, we recommend human approval rather than automatic application.

\clearpage
\section{Alignment Judge Prompt}
\label{app:judge-prompt}

To classify whether a \benchguard{} finding aligns with a human expert revision, we use an LLM judge that receives each (finding, gold issue) pair and classifies the alignment as \emph{aligned}, \emph{partial}, or \emph{unrelated}.
The judge is called once per (gold issue, finding) pair; results are cached to disk to avoid redundant calls.

\paragraph{System Prompt.}

\begin{promptbox}{Alignment Judge --- System Prompt}
You compare two issue descriptions and decide: would a human reviewer say these point to the same underlying problem?

\medskip
Ignore differences in wording, detail level, abstraction, or framing. If fixing one would also fix (or make irrelevant) the other, they describe the same problem.

\medskip
\textbf{Verdicts:}
\begin{itemize}[nosep,leftmargin=*]
  \item \textbf{ALIGNED}: same problem --- a single fix resolves both
  \item \textbf{PARTIAL}: flags the correct functional area but does not pinpoint the exact issue --- a coarser identification of the same underlying problem
  \item \textbf{UNRELATED}: different problems entirely
\end{itemize}

\medskip
When in doubt between ALIGNED and PARTIAL, choose ALIGNED if the root cause is shared.
When in doubt between PARTIAL and UNRELATED, choose PARTIAL.

\medskip
Respond with JSON only: \texttt{\{"verdict": "ALIGNED|PARTIAL|UNRELATED", "reasoning": "one sentence"\}}
\end{promptbox}

\paragraph{User Prompt Template.}
Each pair is presented with the gold issue's description and evidence alongside the finding's title, description, and confidence score.

\begin{promptbox}{Alignment Judge --- User Prompt Template}
\texttt{\#\# Gold Issue}\\
\texttt{Task:} \{task\_id\}\\
\texttt{Issue:} \{issue\_description\}\\
\texttt{Evidence:} \{issue\_evidence\}

\medskip
\texttt{\#\# BenchGuard Finding}\\
\texttt{Title:} \{finding\_title\}\\
\texttt{Description:} \{finding\_description\}\\
\texttt{Confidence:} \{finding\_confidence\}

\medskip
Do these describe the same problem?
\end{promptbox}

\end{document}